\documentclass{article}
\usepackage{spconf,amsmath,epsfig}
\usepackage{pifont, multirow}
\usepackage{hyperref}
\usepackage{graphicx}
\usepackage{subfig}

\newcommand{\cmark}{\ding{51}}%
\newcommand{\xmark}{\ding{55}}%
\let\OLDthebibliography\thebibliography
\renewcommand\thebibliography[1]{
  \OLDthebibliography{#1}
  \setlength{\parskip}{0pt}
  \setlength{\itemsep}{0pt plus 0.3ex}
}

\makeatletter
\def\bstctlcite{\@ifnextchar[{\@bstctlcite}{\@bstctlcite[@auxout]}}
\def\@bstctlcite[#1]#2{\@bsphack
  \@for\@citeb:=#2\do{%
    \edef\@citeb{\expandafter\@firstofone\@citeb}%
    \if@filesw\immediate\write\csname #1\endcsname{\string\citation{\@citeb}}\fi}%
  \@esphack}
\makeatother

\begin{document}\sloppy
\bstctlcite{IEEEexample:BSTcontrol}
\def\x{{\mathbf x}}
\def\L{{\cal L}}

\title{BRIDGING THE DISTRIBUTION GAP OF VISIBLE-INFRARED PERSON RE-IDENTIFICATION WITH MODALITY BATCH NORMALIZATION}
%
\name{Wenkang Li\textsuperscript{1}, Ke Qi\textsuperscript{1}\thanks{Corresponding author: Ke Qi (qikersa@163.com)}, Wenbin Chen\textsuperscript{1}, Yicong Zhou\textsuperscript{2} \thanks{This work was supported by Science Foundation of Guangdong Province under grant No. 2017A030313374.}}
\address{\ninept \textsuperscript{1}School of Computer Science and Cyber Engineering, Guangzhou University, Guangzhou, China \\
\ninept \textsuperscript{2}Department of Computer and Information Science,
University of Macau, Taipa, Macau
}

\maketitle

\begin{abstract}
Visible-infrared cross-modality person re-identification (VI-ReID), whose aim is to match person images between visible and infrared modality, is a challenging cross-modality image retrieval task. Most existing works integrate batch normalization layers into their neural network, but we found out that batch normalization layers would lead to two types of distribution gap: 1) inter-mini-batch distribution gap---the distribution gap of the same modality between each mini-batch; 2) intra-mini-batch modality distribution gap---the distribution gap of different modality within the same mini-batch. To address these problems, we propose a new batch normalization layer called Modality Batch Normalization (MBN), which normalizes each modality sub-mini-batch respectively instead of the whole mini-batch, and can reduce these distribution gap significantly. Extensive experiments show that our MBN is able to boost the performance of VI-ReID models, even with different datasets, backbones and losses.
\end{abstract}
\section{Introduction}
\label{sec:intro}

Person re-identification is an image retrieval task, which matches person images across multiple disjoint cameras. Person re-identification plays an important role in the security field, because these cameras are usually deployed in different locations, and the results of person re-identification can help track the suspects.

\begin{figure}[t]
\includegraphics[width=8.5cm]{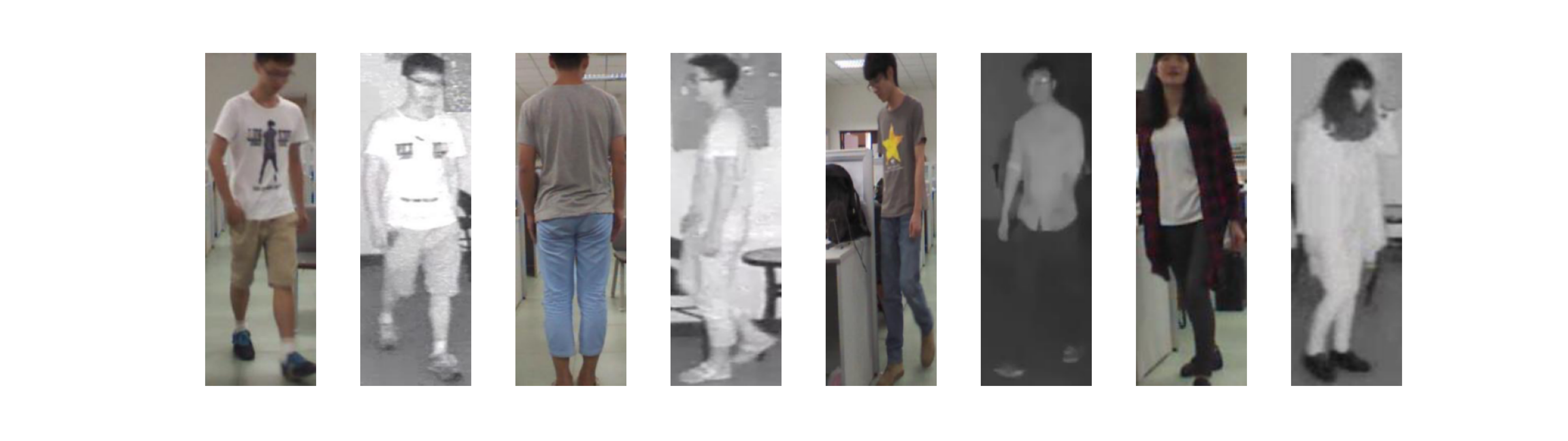}
\caption{\ninept Example images of SYSU-MM01~\cite{sysumm01} dataset. Images of visible modality and infrared modality are RGB and grayscale images respectively.}
\label{fig:example}
\end{figure}

In recent years, person re-identification between visible cameras has made great progress and achieved surpassing human performance on Market1501 dataset~\cite{survey}. However, visible cameras have poor imaging quality at night, so a lot of cameras switch to infrared mode at night. Therefore, the task of person re-identification between day and night becomes the task of person re-identification between visible and infrared. As shown in Figure~\ref{fig:example}, the differences between visible images and infrared images is that infrared images are grayscale images with more noise and less details. Due to such a huge difference, the existing visible-visible person re-identification model performs poorly on the visible-infrared person re-identification task \cite{survey}. In order to get better day-night person re-identification results, it is necessary to redesign models for the visible-infrared person re-identification task.

\begin{figure}[t]
\centering
\subfloat[batch1 whole norm]{
\begin{minipage}[t]{0.48\linewidth}
  \centering
  \includegraphics[width=4.0cm]{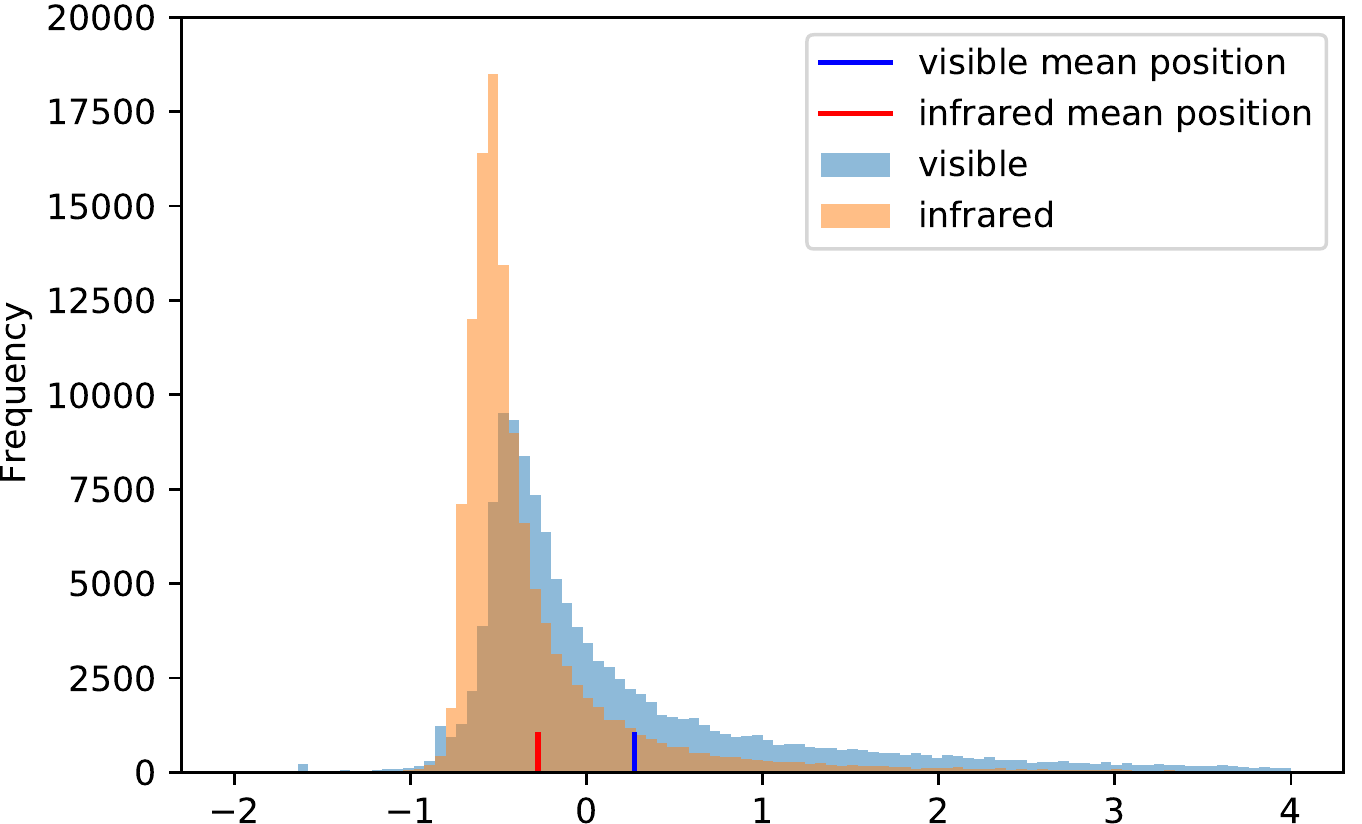}
\end{minipage}
}
\subfloat[batch1 modality norm]{
\begin{minipage}[t]{0.48\linewidth}
  \centering
  \includegraphics[width=4.0cm]{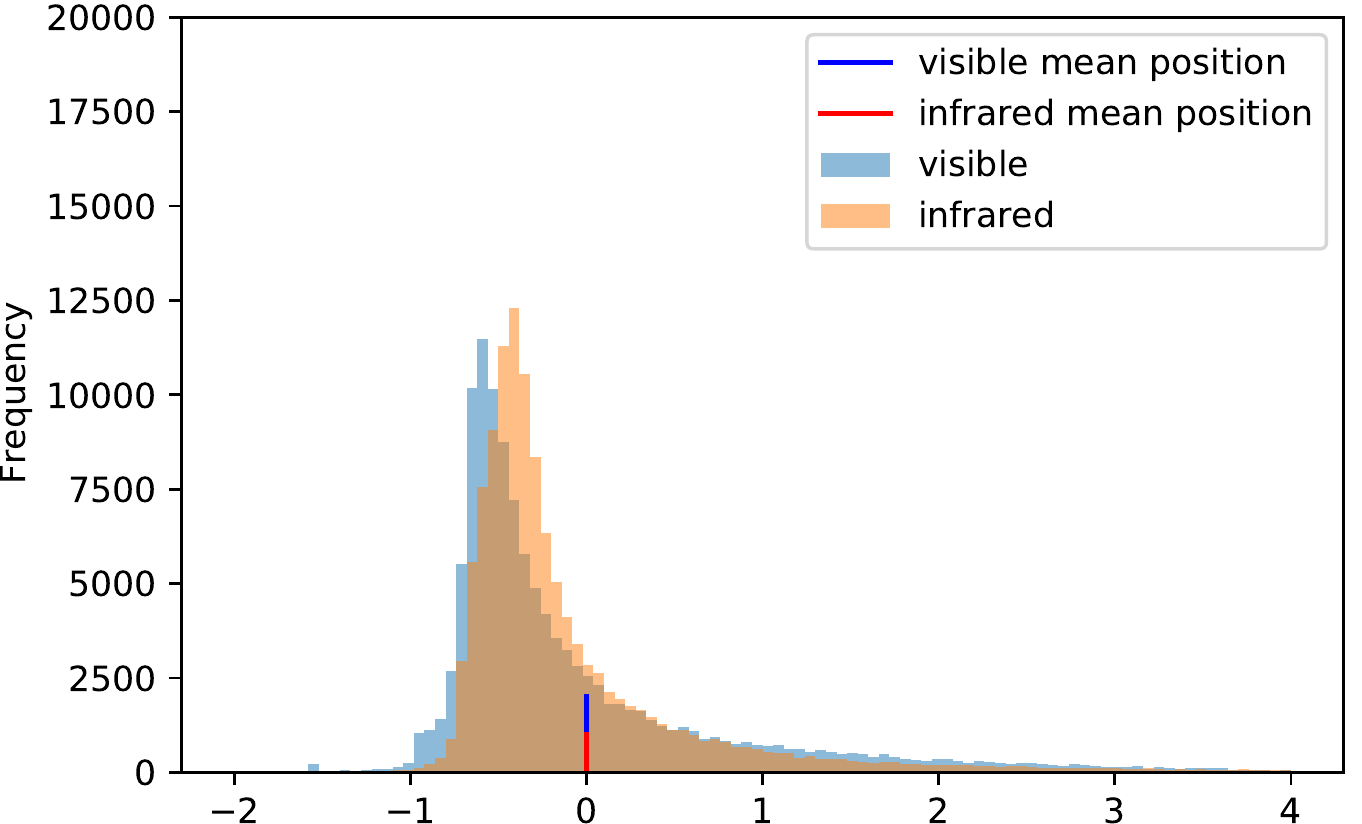}
\end{minipage}
} \\
\subfloat[batch2 whole norm]{
\begin{minipage}[t]{0.48\linewidth}
  \centering
  \includegraphics[width=4.0cm]{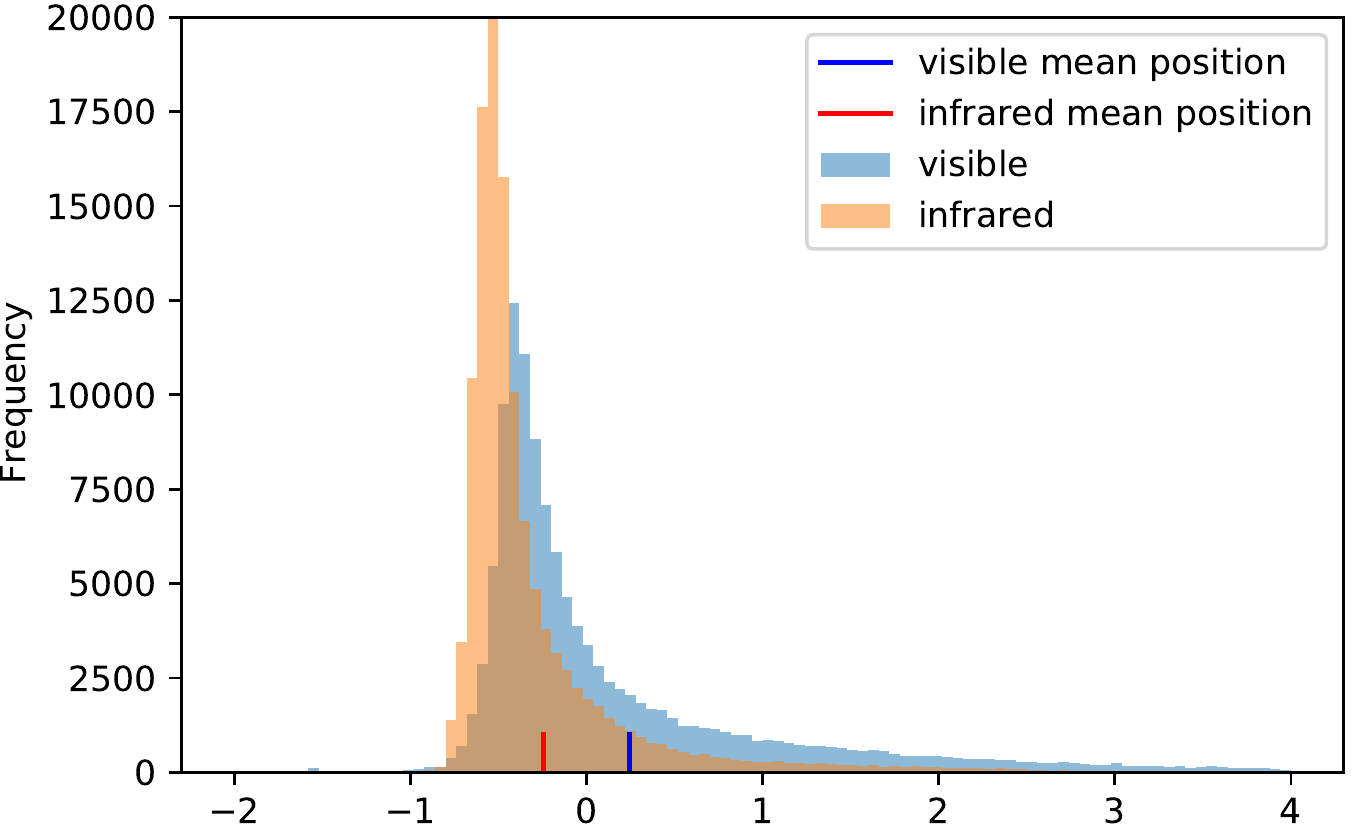}
\end{minipage}
}
\subfloat[batch2 modality norm]{
\begin{minipage}[t]{0.48\linewidth}
  \centering
  \includegraphics[width=4.0cm]{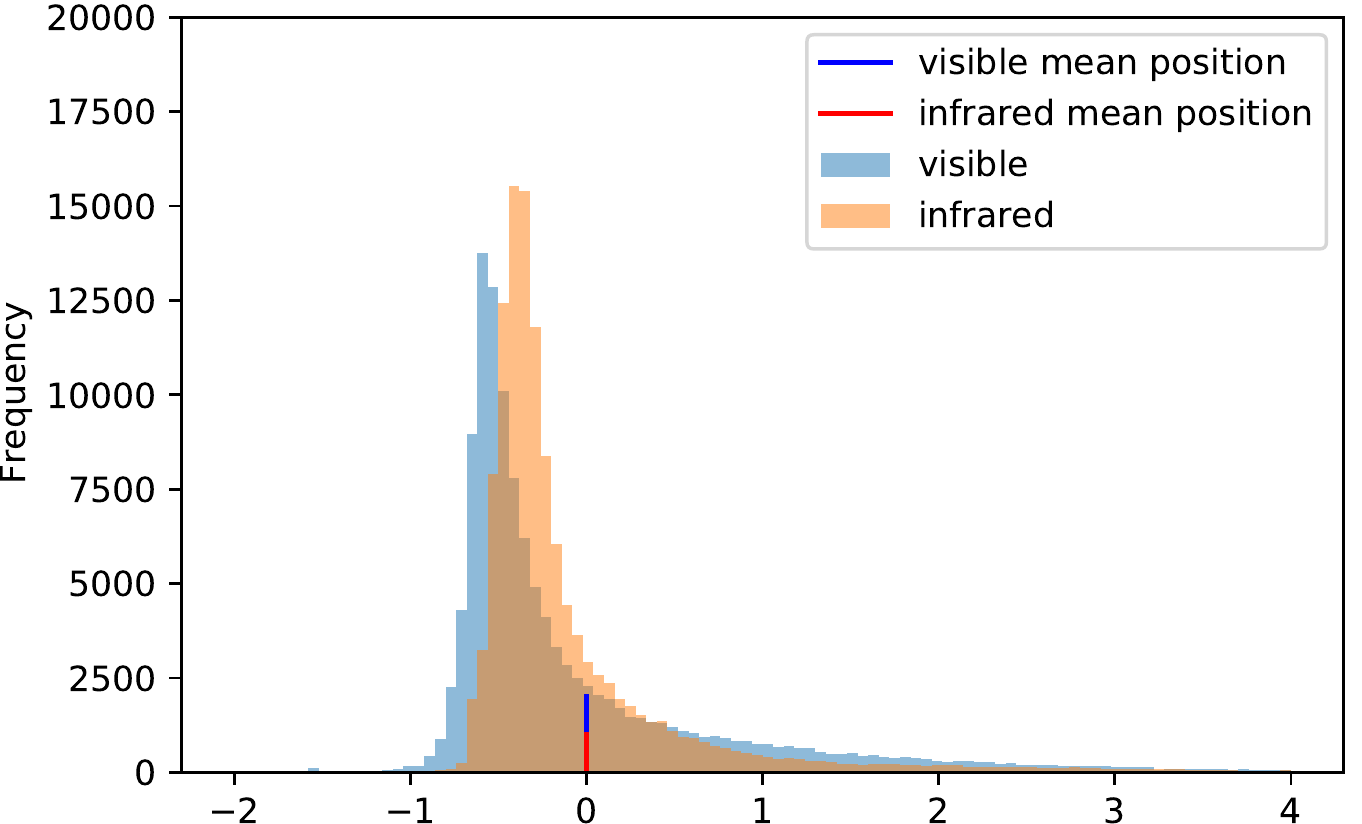}
\end{minipage}
}  \\
\subfloat[batch3 whole norm]{
\begin{minipage}[t]{0.48\linewidth}
  \centering
  \includegraphics[width=4.0cm]{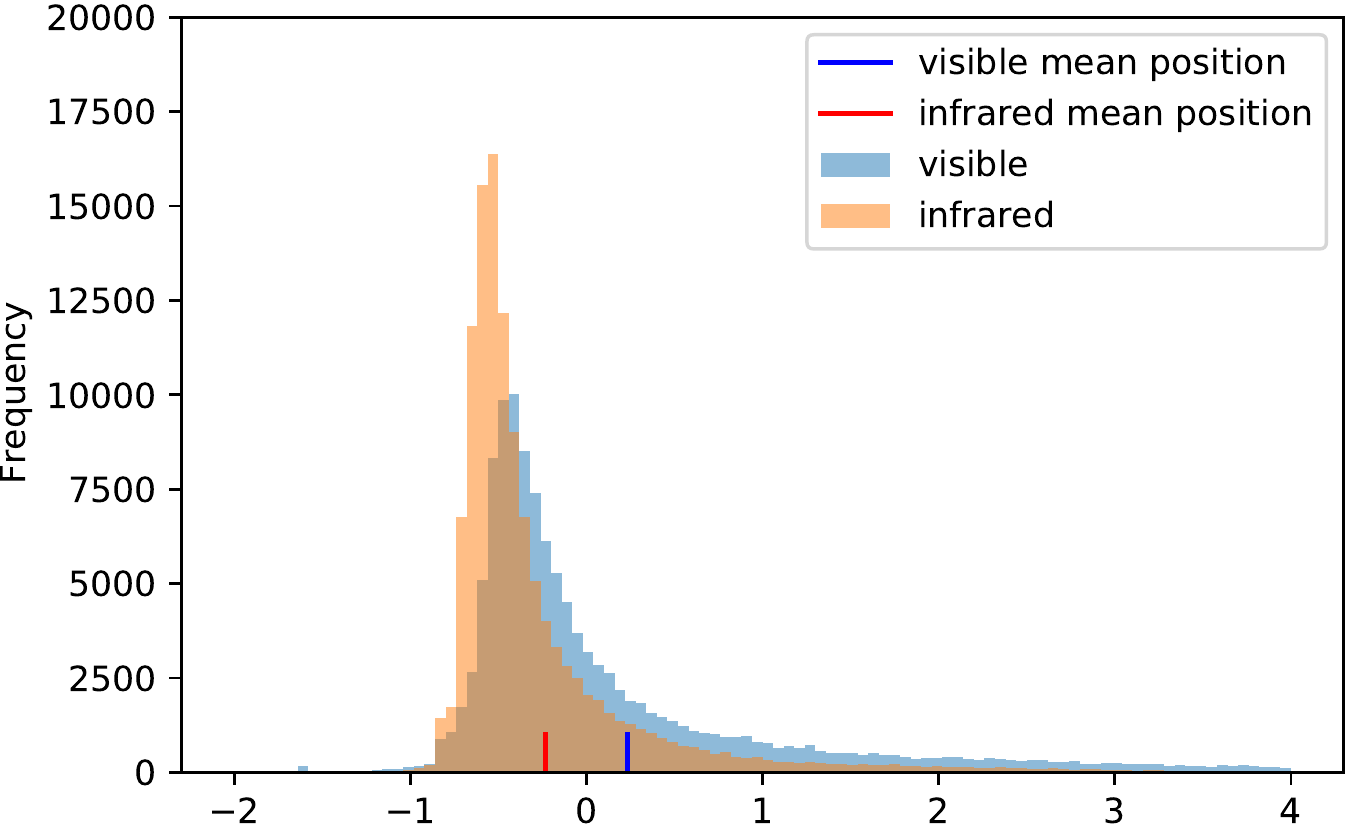}
\end{minipage}
}
\subfloat[batch3 modality norm]{
\begin{minipage}[t]{0.48\linewidth}
  \centering
  \includegraphics[width=4.0cm]{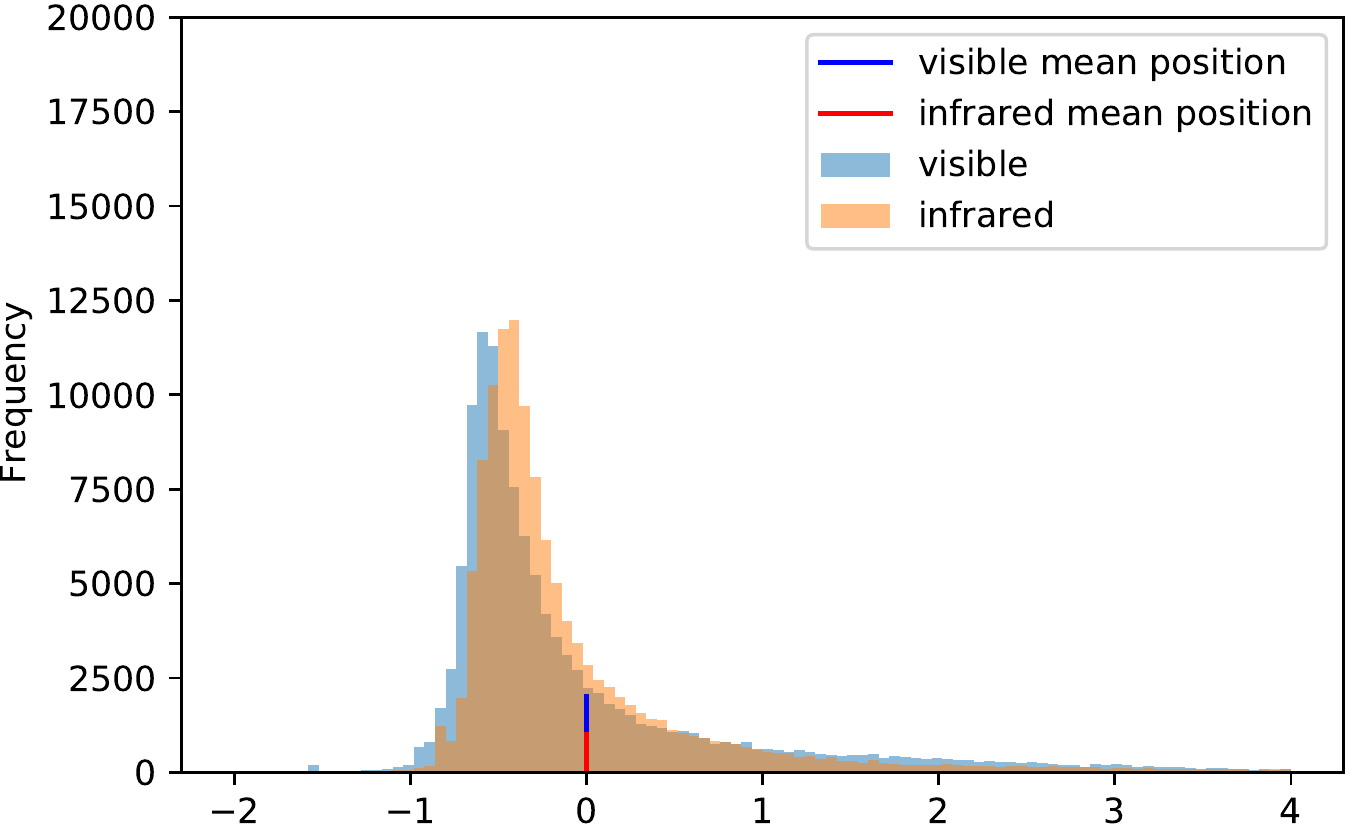}
\end{minipage}
}    
\caption{\ninept Illustration of the histogram of two different normalization methods. We randomly selected 3 different mini-batches, fed them into the BN baseline model, got the outputs of the first channel of the Resnet50 stage1, normalized these outputs with the two normalization methods respectively. (a)(c)(e) are the histograms of the results of whole mini-batch normalization, which is employed by Batch Normalization. (b)(d)(f) are the histograms of the results of modality sub-batch normalization, which is employed by our Modality Batch Normalization.}
\label{fig:distribution}
\end{figure}

\begin{figure}[t]
\centering
\subfloat[statistics of the whole mini-batch normalization]{\label{gap_all}
\begin{minipage}[b]{1.0\linewidth}
  \centering
  \includegraphics[width=8.0cm]{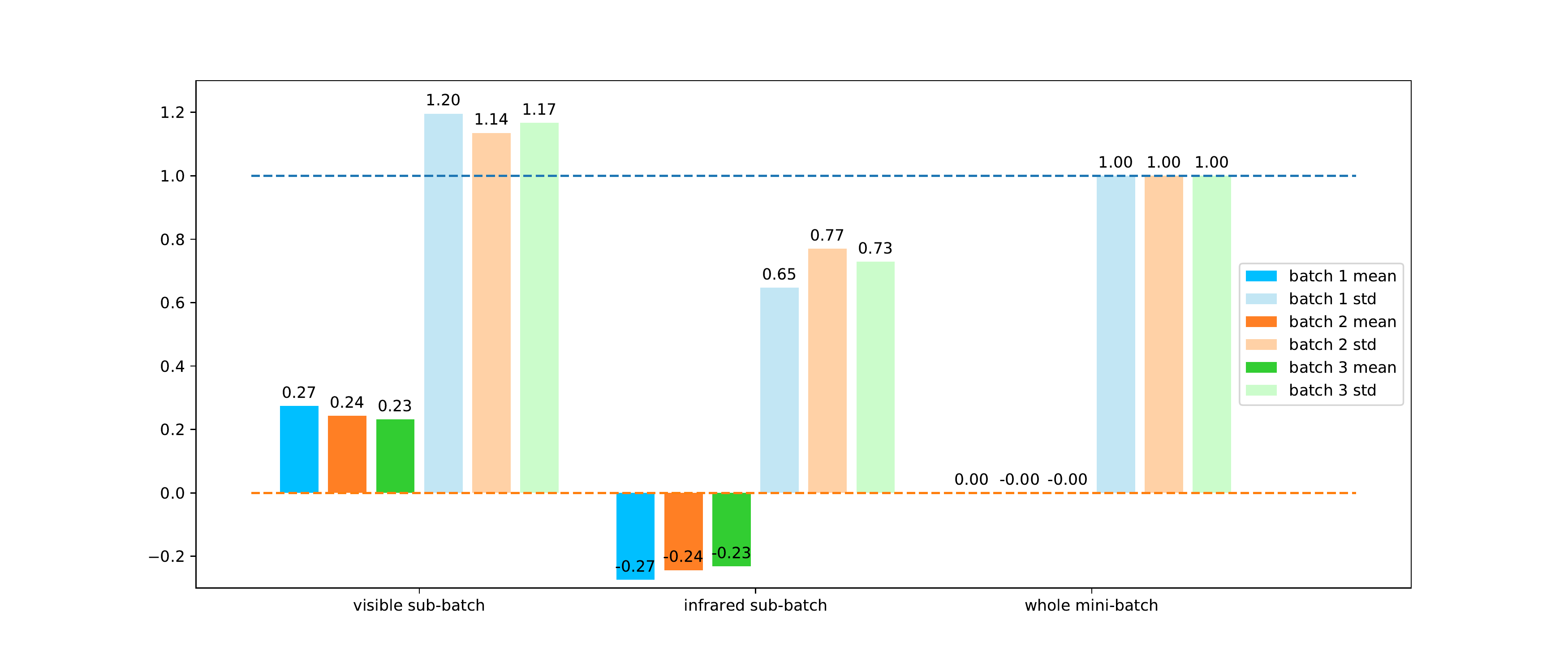}
\end{minipage}
} \\
\subfloat[statistics of the modality sub-mini-batch normalization]{\label{gap_self}
\begin{minipage}[b]{1.0\linewidth}
  \centering
  \includegraphics[width=8.0cm]{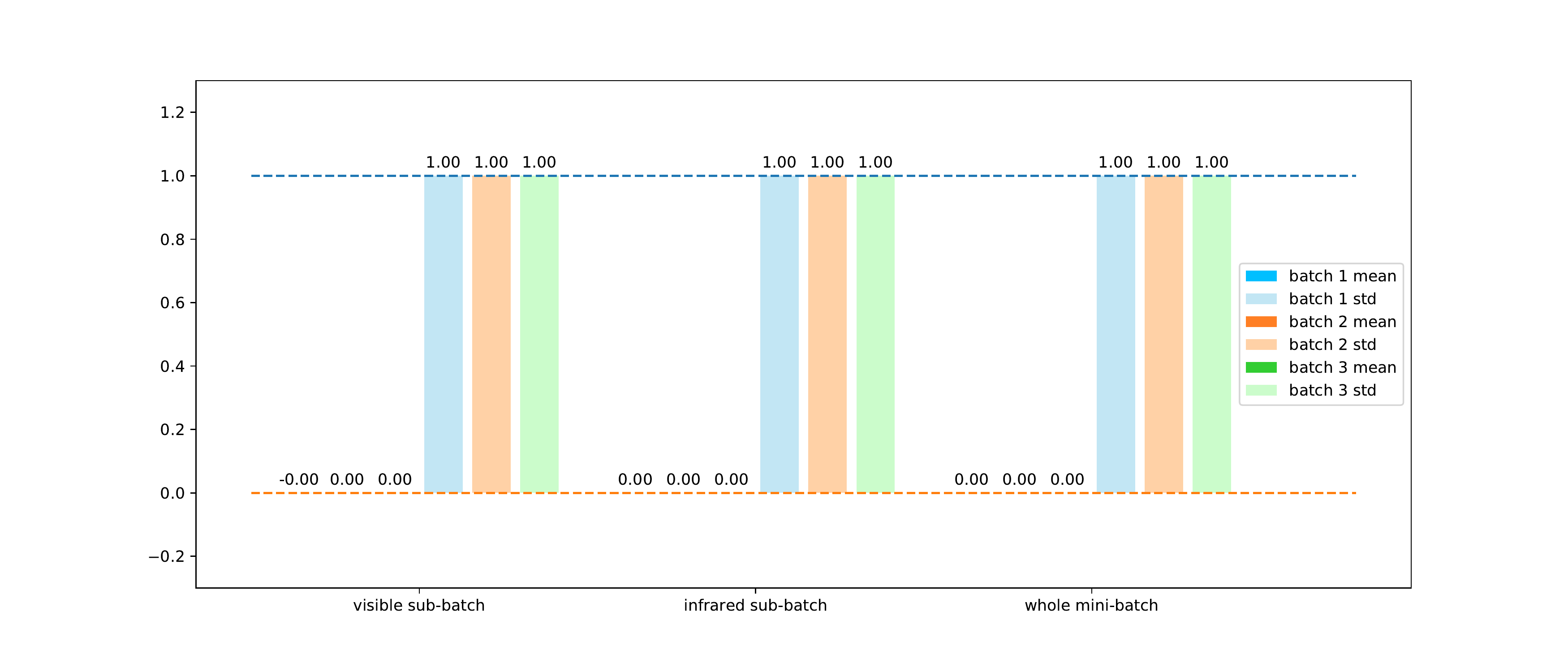}
\end{minipage}
}
\caption{\ninept  Statistics of the same three normalized batches in Figure \ref{fig:distribution}.}
\label{fig:gap}
\end{figure}

To balance the number of training samples, most existing models adopt the 2PK sampling strategy, which first randomly selects P persons, and then randomly selects K visible images and K infrared images of each selected person. It means that each mini-batch contains the same number of visible images and infrared images during the training phase. Moreover, these models integrate Batch Normalization\cite{bn} layers into their neural network, so they normalize the whole mini-batch containing images of different modality. As shown in Figure~\ref{fig:distribution} and \ref{fig:gap}, we found out that this setting will lead to two types of distribution gap: 1) Inter-mini-batch distribution gap. For the same modality, we can observe that the mean and standard deviation between different mini-batches of that modality are quite different; 2) Intra-mini-batch modality distribution gap. Within the same mini-batch, we can observe that the mean and standard deviation of different modality are quite different. We believe that these distribution gaps will harm the performance of models, so we propose a new batch normalization layer called Modality Batch Normalization (MBN), which normalizes each modality sub-mini-batch respectively instead of the whole mini-batch. Comparing Figure~\ref{gap_all} with Figure~\ref{gap_self}, which applied the whole mini-batch normalization and modality sub-batch normalization respectively, we can find out that there is no distribution gap existing in the latter one. To demonstrate the effectiveness of our MBN, we simply replace the BN of existing models with MBN, and extensive experiments show that our MBN is able to boost the performance of VI-ReID models, even with different datasets, backbones and losses.

Our main contributions are summarized as follows:
\begin{itemize}
  \item We found the distribution gaps caused by batch normalization and designed a new batch normalization layer called Modality Batch Normalization (MBN) to deal with this problem.
  \item Extensive experiments show that our MBN is able to boost the performance of VI-ReID models by simply replacing the BN with MBN.
  \item We establish a strong baseline for VI-ReID, which is so simple that will not conflict with most other methods, such as partial features, attention mechanisms, etc.
\end{itemize} 

\section{RELATED WORK}
The basic solution of person re-identification is that, maps each person image into a feature embedding vector, then compute the cosine or euclidean distance between vectors as the similarity between images. For single modality person re-identification, BOT \cite{bot} establishs a strong baseline model only using global features. MGN \cite{mgn} splits the output feature maps into multiple granularities and learns local features for each of them. AlignedReID \cite{alignedreid} aligns local features between different images.  ABD-Net \cite{abdnet} proposes a attention mechanism to enhance important areas or channels in the feature maps.

In addition to dealing with the common problems of person re-identification, visible-infrared person re-identification also needs to deal with the problems caused by modality differences. Some existing works addressed this by GAN-base methods. AlignGAN~\cite{aligngan} aligns pixels and features at the same time. CmGAN~\cite{cmgan} only uses adversarial learning to make the features of the two modalities indistinguishable. X Modality~\cite{xmodality} introduces an intermediate modality. Some research is about feature learning. EDFL~\cite{edfl} enhances the discriminative feature learning; MSR~\cite{msr} learns modality-specific representations. Some other works focus on metric learning. BDTR~\cite{bdtr} calculates the triplet loss of intra-modality and inter-modality respectively; HPILN~\cite{hpiln} calculates the triplet loss of inter-modality in addition to the global triplet loss; HC~\cite{hc} shortens the Euclidean distance between the two modality centers. Recently, AGW~\cite{survey} adopts a attention mechanism and DDAG~\cite{ddag} use graph neural networks to generate more useful features.

\section{PROPOSED METHOD}

\begin{figure}[t]
\includegraphics[width=8.5cm]{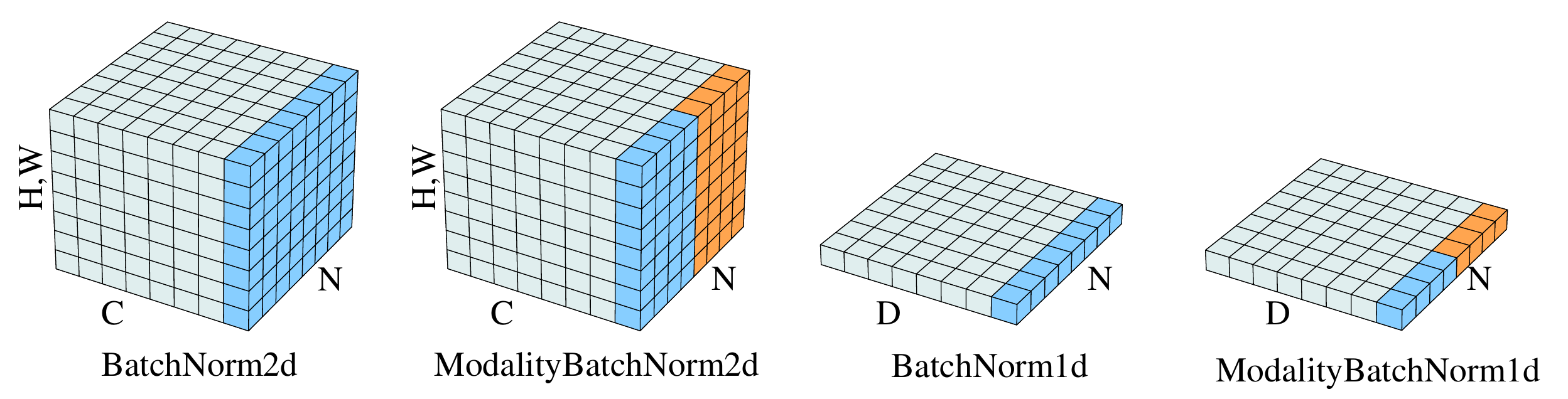}
\caption{\ninept Illustration of normalization methods. Each subplot shows a feature map tensor, with N as the batch axis, C/D as the channel/dimension axis, and (H, W) as the spatial axes. The pixels of the same dark color are normalized by the same mean and variance, computed by aggregating the values of these pixels. Batch Normalization normalizes the whole mini-batch, while Modality Batch Normalization normalizes each modality sub-mini-batch.}
\label{fig:cube}
\end{figure}

\subsection{Batch normalization and distribution gaps}

Batch normalization \cite{bn} was proposed to reduce the internal covariate shift. BN first normalizes the values within the whole mini-batch for each channel, as illustrated in Figure~\ref{fig:cube}, then linearly transforms them with learnable parameters $\gamma$ and $\beta$. Given a value $x\in R^{N\times C\times H\times W}$ belonging to  the input feature map where N is the batch size, C the is channel size, H is the height and W is the width, BN can be expressed as:
\begin{equation}
{\hat{x}}_{n,c,h,w}=BN\left(x_{n,c,h,w}\right)=\gamma_c{\dot{x}}_{n,c,h,w}+\beta_c 
\end{equation}
$\gamma_c$ and $\beta_c$ are learnable parameters of each channel and
\begin{equation}
{\dot{x}}_{n,c,h,w}=\frac{x_{n,c,h,w}-\mu_c}{\sqrt{\sigma_c^2+\epsilon\ \ }}
\end{equation}
$\epsilon$ is a small constant value to avoid divide-by-zero, $\mu_c$ and $\sigma_c^2$ are computed by:
\begin{equation}
\mu_c=\frac{\sum_{n=1}^{N}\sum_{h=1}^{H}\sum_{w=1}^{W}x_{n,c,h,w}}{N\ast H\ast W}
\end{equation}
\begin{equation}
\sigma_c^2=\frac{\sum_{n=1}^{N}\sum_{h=1}^{H}\sum_{w=1}^{W}\left(x_{n,c,h,w}-\mu_c\right)^2}{N\ast H\ast W}
\end{equation}
In the test phase, the batch size of the input may be 1, which means that computing the $\mu_c$ and $\sigma_c^2$ is useless. To deal with this, BN uses moving average  ${\bar{\mu}}_c$ and ${\bar{\sigma}}_c^2$ recorded during training phase, which are computed by:
\begin{equation}
\mu_c^{(t+1)}=\left(1-\alpha\right){\bar{\mu}}_c^{\left(t\right)}+\alpha\mu_c^{\left(t\right)}
\end{equation}
\begin{equation}
\left(\sigma_c^{\left(t+1\right)}\right)^2=\left(1-\alpha\right)\left({\bar{\sigma}}_c^{\left(t\right)}\right)^2+\alpha\left(\sigma_c^{\left(t\right)}\right)^2
\end{equation}
$\alpha$ is the momentum factor and $t$ represents the $t^{th}$ mini-batch.

The intuition behind BN is that, the importance of each channel of the feature maps should be determined by the network, not by the input itself. So BN makes each channel distributed around zero, then learns to scale and shift each channel. However, the whole mini-batch normalization method employed by BN is not suitable for visible-infrared person re-identification, because it will lead to two types of distribution gap. As shown in Figure~\ref{gap_all}, the mean and standard deviation of different modality sub-batches within the same mini-batch are quite different, though the whole mini-batch has already been zero mean and unit standard deviation. That is the intra-mini-batch modality distribution gap. Also shown in Figure~\ref{gap_all}, the mean and standard deviation of the same modality between different mini-batches are quite different. That is the inter-mini-batch distribution gap. The intra-mini-batch modality distribution gap is a quite strong assumption provided by inputs, we argue that it should be determined by the network rather than the input itself, just like the intuition behind BN. What’s more, even if such a distribution gap is beneficial to the network, the inter-mini-batch distribution gap shows that it’s fluctuating. To deal with these, we propose a new batch normalization layer called Modality Batch Normalization (MBN).

\subsection{Modality batch normalization}

Since the whole mini-batch normalization will lead to the two types of distribution gap, we normalize each modality sub-mini-batch respectively, as illustrated in Figure~\ref{fig:cube}. Assuming that $V$, $I$ includes all the visible samples and infrared samples within the mini-batch respectively, we denote $M\in\{V, I\}$.

the mean $\mu_{M,\ c}$ and the variation $\sigma_{M,c}^2$ of each channel belonging to each modality are computed by:
\begin{equation}
\label{eq:mu}
\mu_{M,c}=\frac{\sum_{n\in M}\sum_{h=1}^{H}\sum_{w=1}^{W}x_{n,c,h,w}}{M\ast H\ast W}
\end{equation}
\begin{equation}
\sigma_{M,c}^2=\frac{\sum_{n\in M}\sum_{h=1}^{H}\sum_{w=1}^{W}\left(x_{n,c,h,w}-\mu_{M,c}\right)^2}{M\ast H\ast W}
\end{equation}
So the normalized values are computed by: 
\begin{equation}
{\dot{x}}_{n,c,h,w}= \begin{cases}
\frac{x_{n,c,h,w}-\mu_{V,c}}{\sqrt{\sigma_{V,c}^2+\epsilon\ \ }}\ \ &n\in V \\
\frac{x_{n,c,h,w}-\mu_{I,c}}{\sqrt{\sigma_{I,c}^2+\epsilon\ \ }}\ \ &n\in I
\end{cases}
\end{equation}
We record the moving average ${\bar{\mu}}_{M,c}$ and ${\bar{\sigma}}_{M,c}^2$ for each modality:
\begin{equation}
\mu_{M,c}^{(t+1)}=\left(1-\alpha\right){\bar{\mu}}_{M,c}^{\left(t\right)}+\alpha\mu_{M,c}^{\left(t\right)}
\end{equation}
\begin{equation}
\left(\sigma_{M,c}^{\left(t+1\right)}\right)^2=\left(1-\alpha\right)\left({\bar{\sigma}}_{M,c}^{\left(t\right)}\right)^2+\alpha\left(\sigma_{M,c}^{\left(t\right)}\right)^2
\end{equation}

Comparing Figure~\ref{gap_all} and Figure~\ref{gap_self}, which applied the whole mini-batch normalization and modality sub-batch normalization respectively, we can find out that there is no distribution gap existing in the latter one.
The last thing to determine is whether we should share learnable affine parameters between modalities. As discussed before, if the modality distribution differences can help the network, we should use modality specific learnable parameters to make it capable of taking advantage of modality differences. But if these modality differences harms, it's hard for network to align two learnable parameters if we don't share these learnable parameters. It is difficult to decide, so we proposed two types of MBN, which are marked as $MBN_{shared}$ and $MBN_{specific}$. The difference between the two is that the former shares learnable affine parameters between modalities, while the latter does not.
\begin{equation}
MBN_{shared}\left(x_{n,c,h,w}\right)=\gamma_c{\dot{x}}_{n,c,h,w}+\beta_c
\end{equation}
\begin{equation}
MBN_{specific}\left(x_{n,c,h,w}\right)=  \begin{cases}
\gamma_{V,c}{\dot{x}}_{n,c,h,w}+\beta_{V,c} &n\in V \\
\gamma_{I,c}{\dot{x}}_{n,c,h,w}+\beta_{I,c} &n\in I
\end{cases}
\end{equation}

\subsection{Model pipeline}

\begin{figure}[t]
\includegraphics[width=8.5cm]{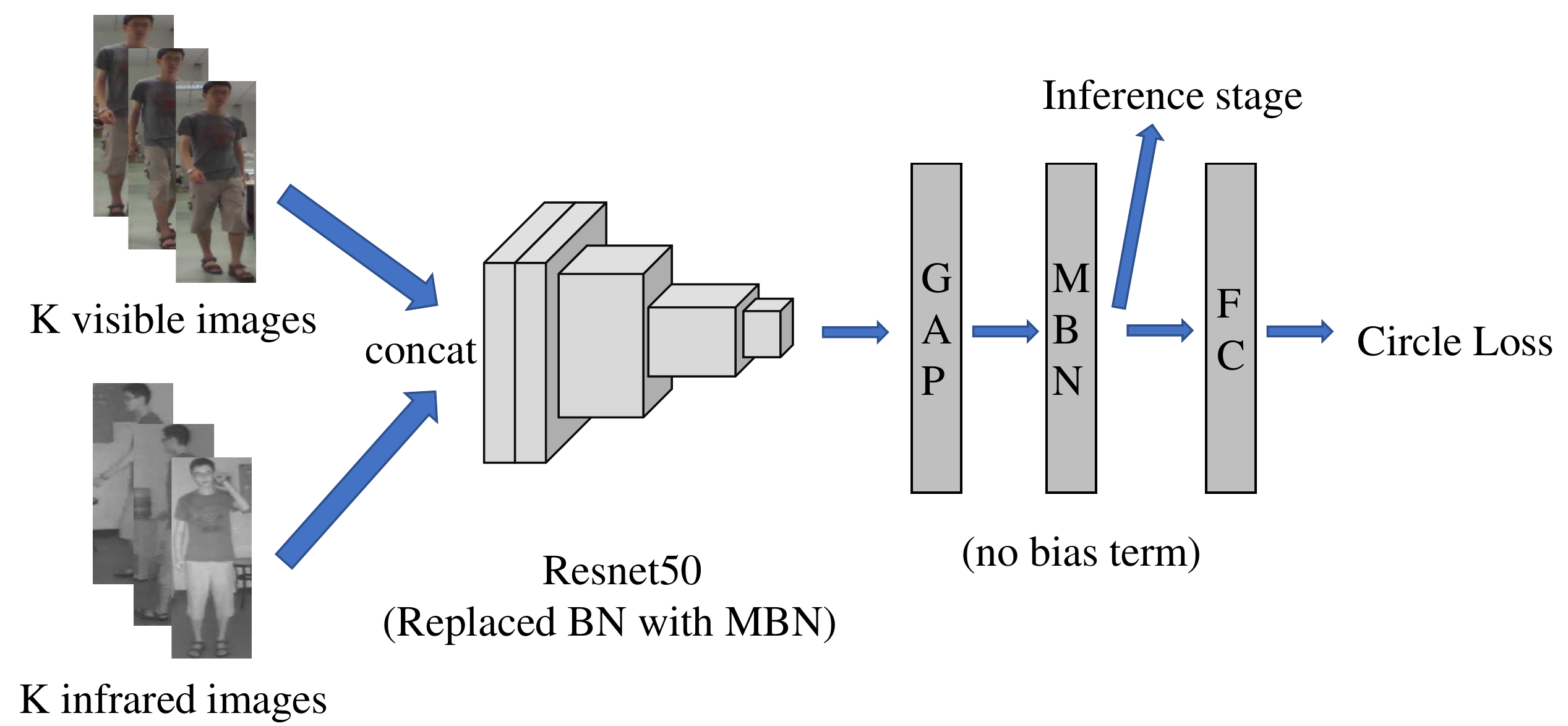}
\caption{\ninept Overall model pipeline. We randomly select K visible and K infrared images to form a mini-batch, then feed these images into the backbone model whose BN are replaced with MBN and get the output feature maps. We use global average pooling to get 1D embedding vector from the output feature maps, and then make it distributed around zero with MBN without bias term. The zero distributed embedding vector is used in inference stage. In training phase, an addition full connect layer is employed to help compute Circle Loss.}
\label{fig:pipeline}
\end{figure}

The overall model pipeline is shown in Figure~\ref{fig:pipeline}. Our model pipeline is modified from BOT~\cite{bot}, which is strong and simple person re-identification baseline model. Comparing with the origin model, we replace all the BN with our MBN, including backbone and head; To keep simple, we use Circle Loss~\cite{circle}, which is a variant of softmax loss, as the loss function instead of softmax loss with triplet loss. Others are kept unchanged. We use cosine value as the similarity metric of embedding vectors.

\section{Experiments}

\subsection{Experiment settings}

We evaluate our methods on SYSU-MM01~\cite{sysumm01} dataset and RegDB~\cite{regdb} dataset. The training set of SYSU-MM01 contains 22258 visible images and 11909 infrared images from 395 IDs. The test set contains 6775 visible images and 3803 infrared images from another 96 IDs. We follow the evaluation protocol of SYSU-MM01, and report the results of all-search one-shot setting. RegDB contains 412 IDs, each ID has 10 visible images, 10 infrared images, a total of 8240 images. We follow the evaluation protocol in Ye et al.~\cite{hdl} for RegDB. We report the CMC and mAP metrics.

\subsection{Implement details}

The backbone containing MBN is initialized with ImageNet pretrained weights. The input images are resized to 320 × 128 for SYSU-MM01 and 256 × 128 for RegDB. Random erasing and random horizontal flip are adopted as data augmentation. We adopt the 2PK sampling strategy, which first randomly selects P persons, and then randomly selects K visible images and K infrared images of each selected person. We set P=6, K=8 for SYSU-MM01 and P=8, K=8 for RegDB. We use the Adam optimizer with lr=6e-4 and wd=5e-4. We warm up 2 epochs and decay the learning rate with 0.1, 0.01 at the 12th epoch and the 16th epoch respectively.
\subsection{Experiment results}

\begin{table}[t]
\begin{center}
{\ninept
\caption{Results of Circle Loss on SYSU-MM01. Backbone and head mean whether they are applied MBN. Rank-1(\%) and mAP(\%) are reported.} \label{tab:circle}
\begin{tabular}{|c|c|c|c|c|}
  \hline
   & backbone & head & rank-1 & mAP   \\
  \hline
  baseline & \xmark & \xmark & 51.0 & 49.2 \\
  \hline
  \multirow{3}{*}{$MBN_{shared}$} & \cmark & \xmark & 56.0	& 54.1  \\
                                  & \xmark & \cmark & 51.9	& 50.5 \\
                                  & \cmark & \cmark & \textbf{56.1}	& \textbf{54.2}  \\
  \hline
  \multirow{3}{*}{$MBN_{specific}$} & \cmark & \xmark & 49.8	& 48.3	 \\
                                    & \xmark & \cmark & 54.1	& 52.5	 \\
                                    & \cmark & \cmark & \textbf{56.3}	& \textbf{54.2}	 \\
  \hline
  mixed & \cmark \par(shared) & \cmark \par(specific) & 55.6	& 53.8  \\
  \hline
\end{tabular}
}
\end{center}
\end{table}

\begin{figure}[t]
\includegraphics[width=8.5cm]{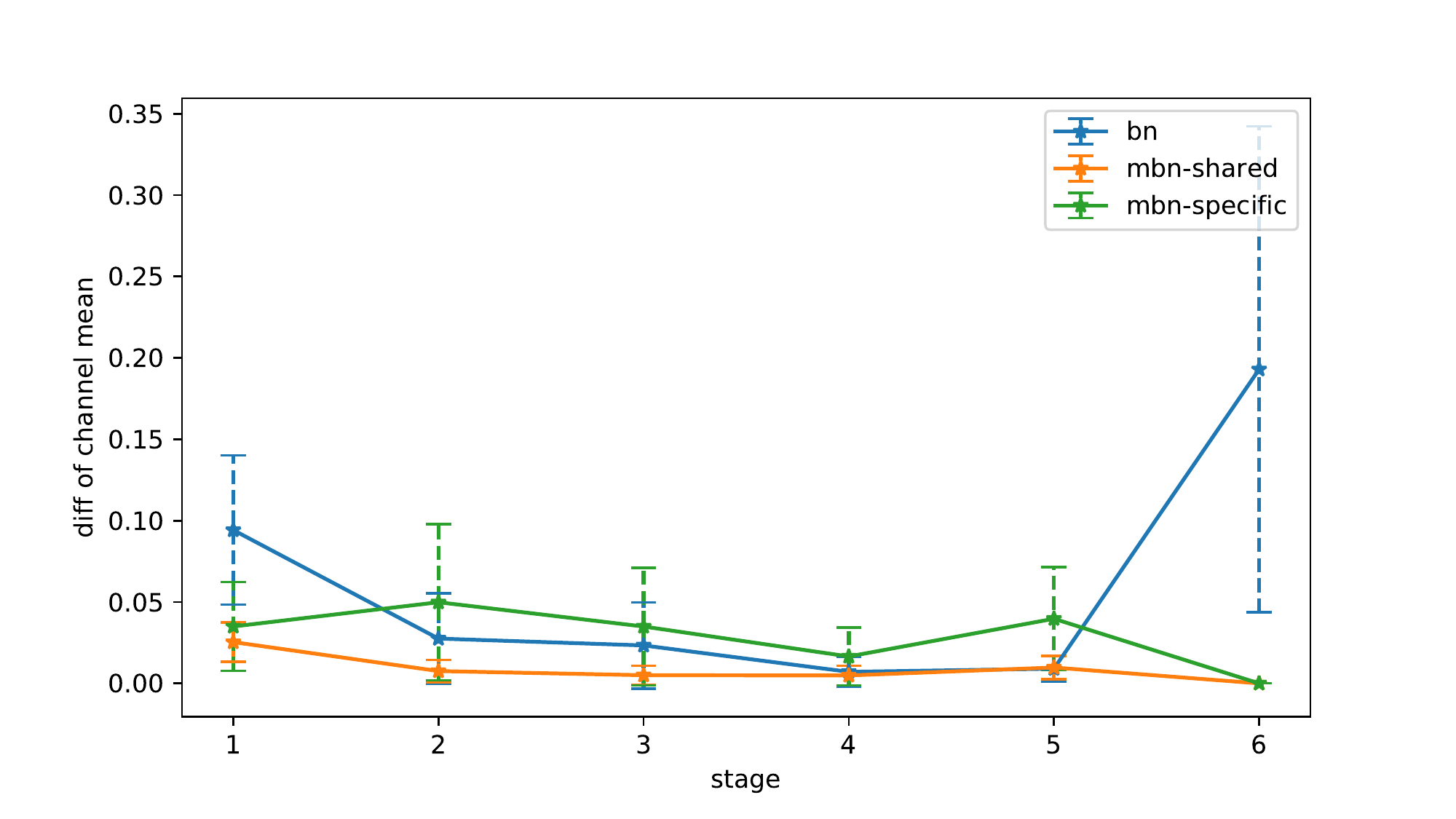}
\caption{\ninept Illustration of intra-mini-batch modality distribution gaps of each stage. It's the statistics of $\left|\mu_{V,c}-\mu_{I,c}\right|$. Stage1-5 are the stages of Resnet50 Backbone, Stage 6 is the output of BN/MBN head.}
\label{fig:chan_mean_diff}
\end{figure}
\begin{figure}[t]
\centering
\subfloat[circle loss]{
\begin{minipage}[b]{0.48\linewidth}
  \centering
  \includegraphics[width=4.0cm]{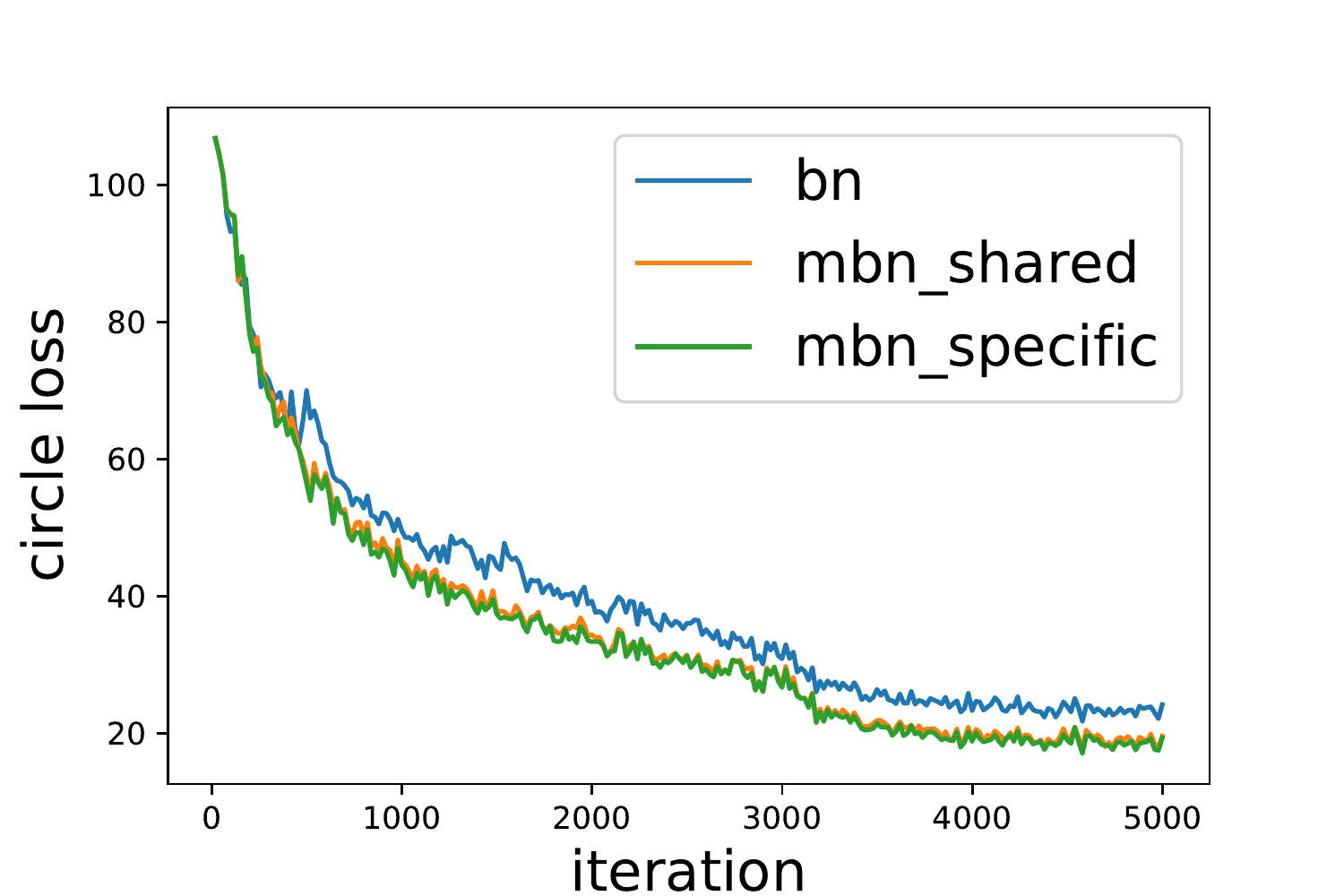}
\end{minipage}
}
\subfloat[mAP]{
\begin{minipage}[b]{0.48\linewidth}
  \centering
  \includegraphics[width=4.0cm]{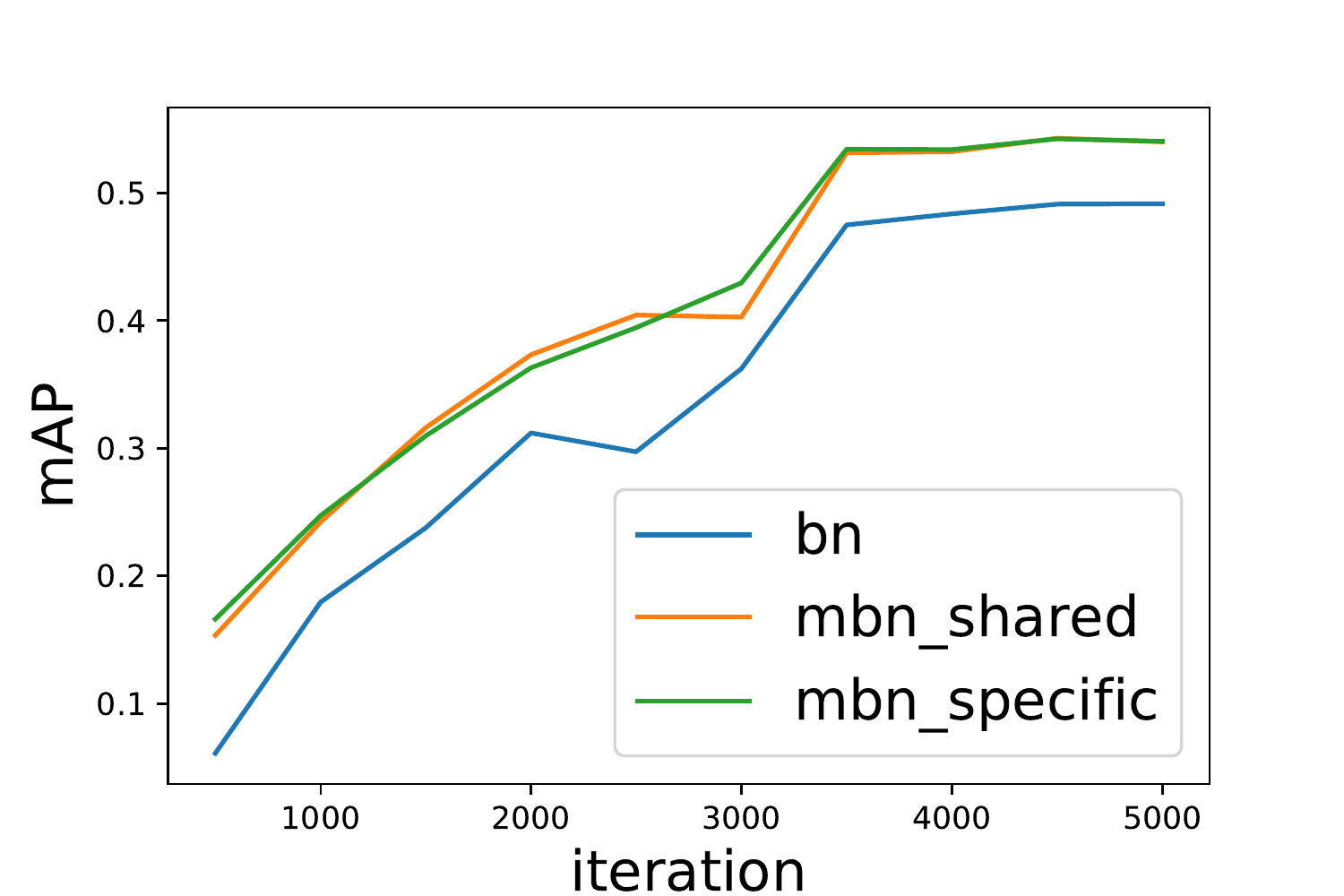}
\end{minipage}
}
\caption{\ninept Training curve of Circle Loss.}
\label{fig:circle_curve}
\end{figure}

\subsubsection{Results of Circle Loss}

The results of Circle Loss are shown in Table~\ref{tab:circle}. We make several observations through this: 1) If $MBN_{shared}$ or $MBN_{specific}$ is applied to the entire model, there can be a 5\% increase on Rank-1 and mAP for the baseline model. 2) Applying $MBN_{shared}$ to the backbone or head alone can improve the performance of the baseline model, but applying it to the backbone alone has a greater performance improvement. 3) Applying $MBN_{specific}$ to backbone alone will reduce performance, while applying it to head alone can improve performance. 4) Mixing $MBN_{shared}$ and $MBN_{specific}$ is no better than using only one of them.

	As shown in Figure~\ref{fig:chan_mean_diff}, we plot the statistics of $\left|\mu_{V,c}-\mu_{I,c}\right|$, the absolute value of channel mean difference of different modalities, which can reflect the intra-mini-batch modality distribution gap. Compared with BN, the modality distribution gap of $MBN_{shared}$ on Backbone is much smaller, while $MBN_{specific}$ has a larger modality distribution gap due to additional affine parameters. This is why applying $MBN_{shared}$ to the backbone alone gets good results but $MBN_{specific}$ get bad results. However, the final MBN on the head, whether it is a shared or specific version, reduces the modality distribution gap to a very low level, which is why the two versions of MBN ultimately have better results.

	Figure~\ref{fig:circle_curve} shows the training curves of Circle Loss. We can see that models with MBN are much easy to fit and always get better performance during training phase.

\subsubsection{Results of softmax loss with triplet loss}

\begin{table}[t]
\begin{center}
{\ninept
\caption{Results of softmax loss with triplet loss on SYSU-MM01. Backbone and head mean whether they are applied MBN. Rank-1(\%) and mAP(\%) are reported.} \label{tab:triplet}
\begin{tabular}{|c|c|c|c|c|}
  \hline
   & backbone & head & rank-1 & mAP   \\
  \hline
  baseline & \xmark & \xmark & 50.2	& 45.7 \\
  \hline
  \multirow{3}{*}{$MBN_{shared}$} & \cmark & \xmark & \textbf{51.1} &	\textbf{47.0}  \\
                                  & \xmark & \cmark & 50.9 &	46.6 \\
                                  & \cmark & \cmark & 51.1	& 46.9  \\
  \hline
  \multirow{3}{*}{$MBN_{specific}$} & \cmark & \xmark & 50.9	& 47.3	 \\
                                    & \xmark & \cmark & 53.3	& 49.9	 \\
                                    & \cmark & \cmark & \textbf{55.3}	& \textbf{52.2}	 \\
  \hline
  mixed & \cmark \par(shared) & \cmark \par(specific) & 54.1	& 50.7  \\
  \hline
\end{tabular}
}
\end{center}
\end{table}

\begin{figure}[t]
\centering
\subfloat[softmax loss]{
\begin{minipage}[b]{0.3\linewidth}
  \centering
  \includegraphics[width=\linewidth]{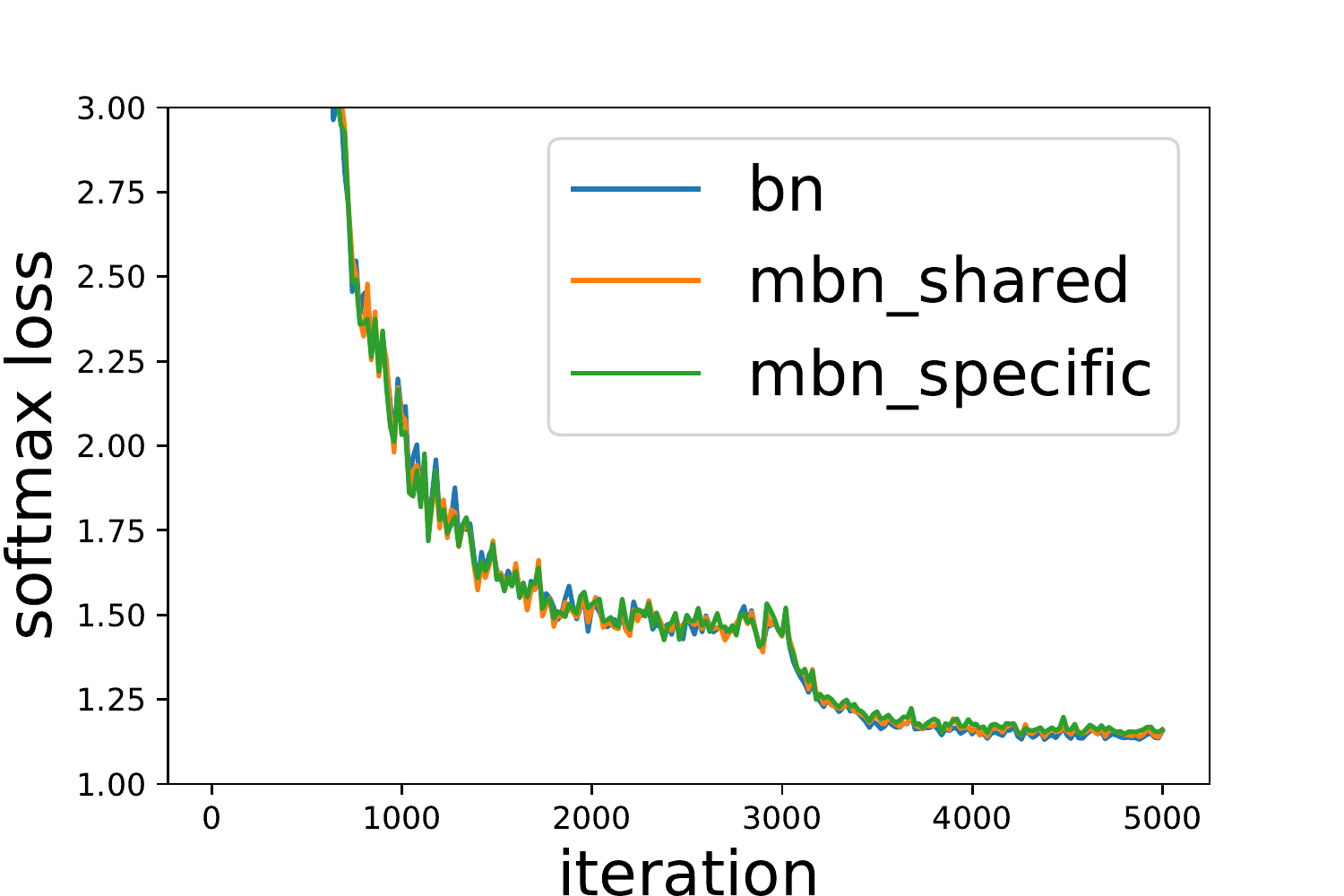}
\end{minipage}
}
\subfloat[triplet loss]{
\begin{minipage}[b]{0.3\linewidth}
  \centering
  \includegraphics[width=\linewidth]{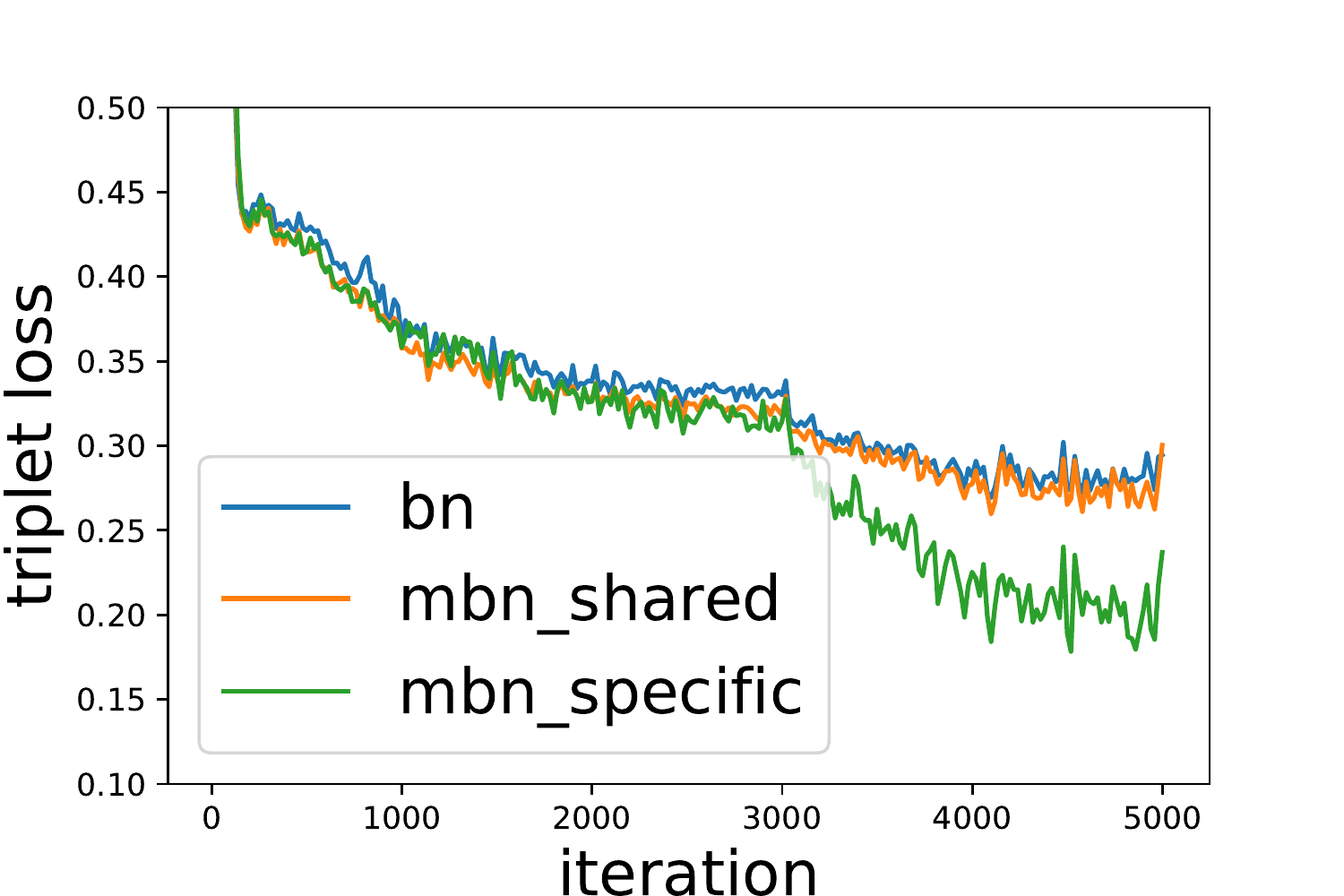}
\end{minipage}
}
\subfloat[mAP]{
\begin{minipage}[b]{0.3\linewidth}
  \centering
  \includegraphics[width=\linewidth]{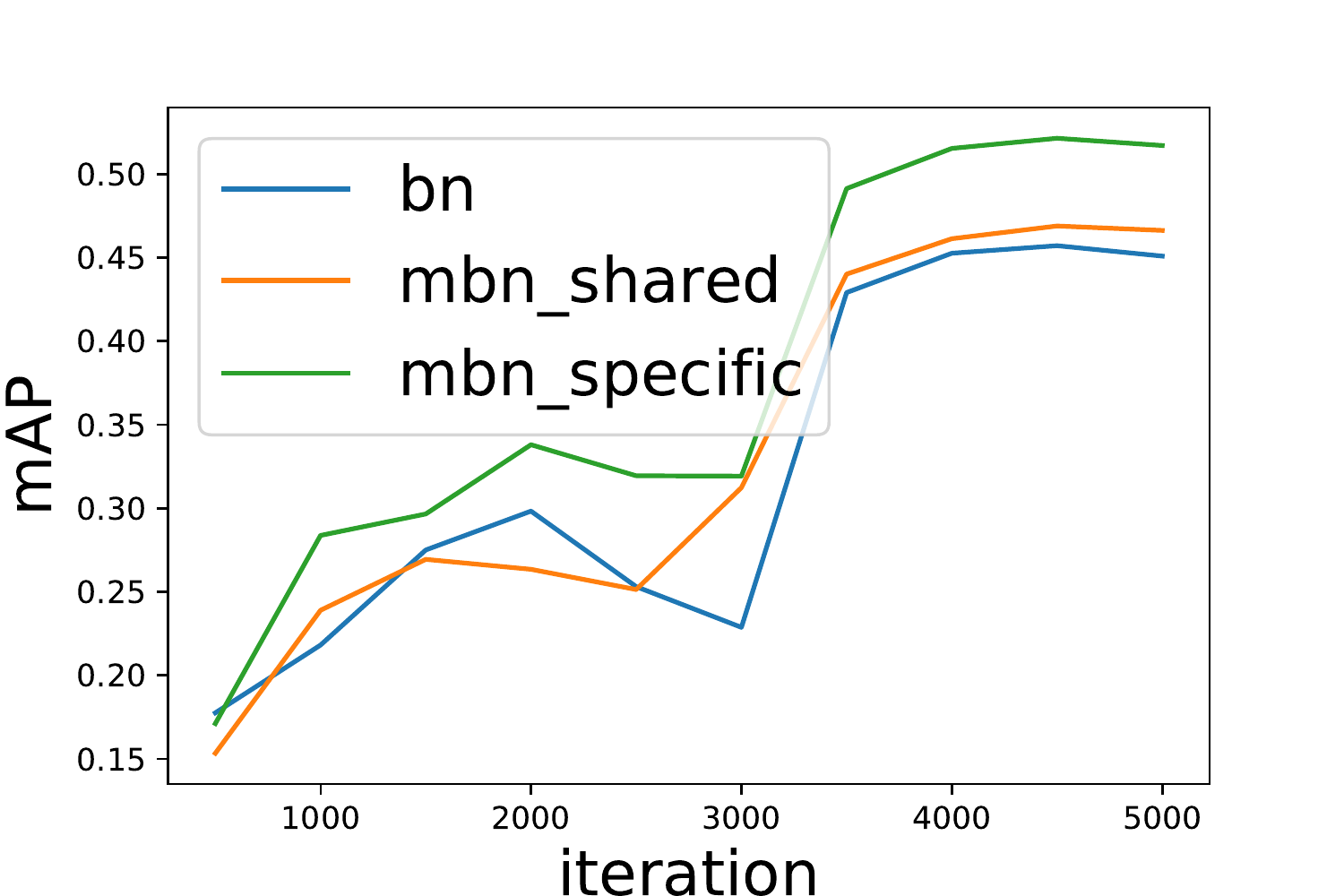}
\end{minipage}
}
\caption{\ninept Training curve of softmax loss with triplet loss..}
\label{fig:triplet_curve}
\end{figure}

We also evaluate our methods with softmax loss with triplet loss, which is used by BOT~\cite{bot}. As shown in Table~\ref{tab:triplet}, most observations are similar to Circle Loss except two: 1) $MBN_{specific}$ gets better performance than $MBN_{shared}$. 2) Applying $MBN_{specific}$ to backbone alone will boost the performance while Circle Loss won’t. As shown in Figure~\ref{fig:triplet_curve}, which plots the training curves of softmax loss with triplet loss, we can see that the softmax loss curves of different models are similar, but the triplet loss curve of the $MBN_{specific}$ model drops faster than the other two models. Therefore, we believe that the reason for the better performance of $MBN_{specific}$ is that triplet loss optimizes the Euclidean distance between samples, so the modality-specific affine parameters in $MBN_{specific}$ are very helpful for optimization.

\subsubsection{Results of Resnext50 backbone and RegDB dataset}

\begin{table}[t]
\begin{center}
{\ninept
\caption{Results of Resnext50~\cite{resnext} with Circle Loss on SYSU-MM01. BN Type is the BN layers used by the entire model. Rank-1(\%) and mAP(\%) are reported.} \label{tab:resnext}
\begin{tabular}{|c|c|c|}
  \hline
  BN type & rank-1 & mAP   \\
  \hline
  BN & 51.7	 & 51.1 \\
  \hline  
  $MBN_{shared}$ & \textbf{53.3} & \textbf{52.4} \\
  $MBN_{specific}$ & 52.4	 & 52.0 \\
  \hline
\end{tabular}
}
\end{center}
\end{table}

\begin{table}[t]
\begin{center}
{\ninept
\caption{Results of Circle Loss on RegDB. BN Type is the BN layers used by the entire model. Rank-1(\%) and mAP(\%) are reported.} \label{tab:regdb}
\begin{tabular}{|c|c|c|c|c|}
  \hline
  \multirow{2}{*}{BN type} & \multicolumn{2}{c|}{Visible to Infrared} &  \multicolumn{2}{c|}{Infrared to Visible} \\ \cline{2-5}

   & rank-1	& mAP & rank-1	& mAP \\
  \hline
  BN & 67.3	& 64.8 & 65.3 & 62.8 \\
  \hline  
  $MBN_{shared}$ & \textbf{67.8}	& \textbf{65.5} & \textbf{66.2} & \textbf{64.2} \\
  $MBN_{specific}$ & 64.7	& 62.8 & 63.6 & 62.1\\
  \hline
\end{tabular}
}
\end{center}
\end{table}

We also evaluate our methods with Resnext50~\cite{resnext} backbone and RegDB~\cite{regdb} dataset, as shown in Table~\ref{tab:resnext}, \ref{tab:regdb} respectively. Due to the lack of GPU memory, for Resnext50, we set P to 5 and K to 8 in the sampling strategy. We can see that both $MBN_{shared}$ and $MBN_{specific}$ can boost the performance of model with Resnext50 backbone and $MBN_{shared}$ get better performance than $MBN_{specific}$. As for the RegDB, $MBN_{shared}$ improves the performance, while $MBN_{specific}$ drops. Considering the small scale of RegDB, we think that the additional affine parameters make $MBN_{specific}$ model overfitting.

\subsubsection{Comparison with state-of-the-art methods}

As shown in Table~\ref{tab:sota}, we compare our methods with state-of-the-art methods. The following observations can be made: 1) With the help of MBN, the Rank-1 and mAP of our model outperform most existing models except HC~\cite{hc}, which employs local features while ours only employ global features. 2) The Rank-10 and Rank-20 are still not as good as SOTA models. It makes sense, because our model  only focuses on resolving modality differences, and don't introduct complex methods such as attention mechanisms to deal with hard cases such as changes in person poses. Therefore, the improvement of hard cases is limited.

\section{conclusion}

In this paper, we propose a new batch normalization layer called modality batch normalization (MBN), which can deal with the distribution gap between different modalities. It significantly boosts the performance of VI-ReID models by simply replacing the BN with MBN. Because the MBN model is very simple, it can be used as a baseline model and be combined with other complex methods to produce a better model. We believe this finding can help researchers develop a better visible-infrared person re-identification model.

\begin{table}[t]
\begin{center}
{\ninept
\caption{Comparison with the state-of-the-arts on SYSU-MM01 dataset with all-search ont-shot setting. Rank-1, Rank-10, Rank-20(\%) and mAP (\%) are reported.} \label{tab:sota}
\begin{tabular}{|c|c|c|c|c|}
  \hline
  method & rank-1 & rank-10 & rank-20 & mAP   \\
  \hline
  cmGAN~\cite{cmgan} & 26.97& 67.51 &80.56 &27.80 \\
  eBDTR~\cite{bdtr} & 27.82 &67.34 &81.34& 28.42 \\
  EDFL~\cite{edfl} & 36.94 &85.42 &93.22 &40.77 \\
  MSR~\cite{msr} & 37.35 & 83.40 &93.34 &38.11 \\
  HPILN~\cite{hpiln} &41.36 &84.78& 94.51 &42.95 \\
  alignGAN~\cite{aligngan} & 42.40 &85.00 &93.70 &40.70 \\
  AGW~\cite{survey} & 47.50 & 84.39 &92.14& 47.65 \\
  X-Modality~\cite{xmodality} & 49.92 &89.79 &95.96& 50.73 \\
  DDAG~\cite{ddag}	& 54.75 &	90.39	& 95.81 &	53.02 \\
  HC~\cite{hc} 	&56.96	&91.50	&96.82&	54.95 \\
  \hline
  baseline (ours) & 50.91&	85.97	&92.88	&49.12 \\  
  $MBN_{shared}$ (ours) & 56.07&88.59	&94.75	&54.28 \\
  \hline
\end{tabular}
}
\end{center}
\end{table}

\bibliographystyle{IEEEtran}
\bibliography{refer}

\end{document}